\documentclass[a4paper,fleqn]{cas-dc}
\usepackage{caption}
\usepackage{array}
\usepackage{algorithm}
\usepackage{algorithmic}
\usepackage{color}
\newcolumntype{C}[1]{>{\centering\arraybackslash}p{#1}}
\usepackage[numbers]{natbib}
\newcommand{\minew}[1]{{\color{black}{#1}}}

\def\tsc#1{\csdef{#1}{\textsc{\lowercase{#1}}\xspace}}
\tsc{WGM}
\tsc{QE}

\begin{document}
\let\WriteBookmarks\relax
\def\floatpagepagefraction{1}
\def\textpagefraction{.001}
\let\printorcid\relax 

\shorttitle{VIFNet: An End-to-end Visible-Infrared Fusion Network for Image Dehazing}    

\shortauthors{Meng Yu et al.}

\title[mode = title]{VIFNet: An End-to-end Visible-Infrared Fusion Network for Image Dehazing}

\tnotemark[1]
\tnotetext[1]{This work is  supported by the National Natural Science Foundation of China under Grant 62003039, 62233002, the CAST program under Grant No. YESS20200126. }

\author[1]{Meng Yu}
\ead{1294033803@qq.com} 

\author[1]{Te Cui}
\ead{cuite1999@bit.edu.cn} 

\author[1]{Haoyang Lu}
\ead{3120230806@bit.edu.cn} 

\author[1]{Yufeng Yue}
\cormark[1] 
\ead{yueyufeng@bit.edu.cn} 

\address[1]{School of Automation, Beijing Institute of Technology, Beijing,  China}

\cortext[1]{Corresponding author} 

\begin{abstract}
\{\color{black}{Image dehazing poses significant challenges in environmental perception. Recent research mainly focus on deep learning-based methods with single modality, while they may result in severe information loss especially in dense-haze scenarios. The infrared image exhibits robustness to the haze, however, existing methods have primarily treated the infrared modality as auxiliary information, failing to fully explore its rich information in dehazing. To address this challenge, the key insight of this study is to design a visible-infrared fusion network for image dehazing. In particular, we propose a multi-scale Deep Structure Feature Extraction (DSFE) module, which incorporates the Channel-Pixel Attention Block (CPAB) to restore more spatial and marginal information within the deep structural features. Additionally, we introduce an inconsistency weighted fusion strategy to merge the two modalities by leveraging the more reliable information. To validate this, we construct a visible-infrared multimodal dataset called AirSim-VID based on the AirSim simulation platform. Extensive experiments performed on challenging real and simulated image datasets demonstrate that VIFNet can outperform many state-of-the-art competing methods.} The code and dataset are available at  https://github.com/mengyu212/VIFNet\_dehazing.
\end{abstract}


\begin{highlights}
\item We propose an end-to-end multimodal fusion dehazing framework to restore high-quality images. \minew{In addition, we provide a visible-infrared dataset for image dehazing based on AirSim, named AirSim-VID, which contains 3 different fog concentration types.}
\item \minew{In the deep feature extraction stage, we present a Deep Structure Feature Extraction (DSFE) module, which incorporates Channel-Pixel Attention Block (CPAB) to explore more spatial and marginal information within the feature maps.}
\item \minew{In the feature weighted fusion stage, an efficient inconsistency fusion strategy is introduced to adjust the fusion weights between two modalities, which emphasizes more reliable and consistent information.}
\end{highlights}


\maketitle

\section{Introduction}
 \minew{Haze is caused by clustered vapors in the air, which scatters light propagation, disrupting the imaging process and reducing image quality.} It is worth noticing that such low-visibility images significantly impact the performance of relevant high-level tasks in autonomous driving, and can even lead to serious accidents. Therefore, image dehazing, aiming to restore a haze-free image from a hazy input, has garnered significant attention during the past few years. Based on the atmospheric scattering theory \cite{b1}, the degradation of image can be mathematically formulated as the following model:   
\begin{equation}I(x)=J(x)t(x)+A(1-t(x)).\label{eq1}\end{equation}
\begin{equation}t(x)=e^{-\beta d(x)}.\label{eq2}\end{equation}
where \emph{I}(\emph{x}) represents the \emph{x}-th pixel of the observed hazy image, and \emph{J}(\emph{x}) is the restored scene radiance, namely, the haze-free image. The transmission map is denoted by \emph{t}(\emph{x}), which is exponentially correlated to scene depth \emph{d}(\emph{x}) and scattering coefficient $\beta$ that reflects the haze density, and \emph{A} is the global atmosphere light.

Following this atmosphere scattering model, the single haze-free image \emph{J}(\emph{x}) can be derived by estimating \emph{t}(\emph{x}) and \emph{A}, separately. Early researchers attempted to remove haze using handcraft priors, including contrast maximization \cite{b2}, dark channel prior (DCP) \cite{b3}, color attenuation prior \cite{b4}, non-local prior \cite{b5}, and haze-lines prior \cite{b6}. \minew{However, these methods only achieved prominent results when the algorithms aligned with particular priors. For example, DCP \cite{b3} struggled to remove haze in sky regions that didn't satisfy dark channel prior.} To relax the above assumptions and improve the robustness of the dehazing algorithms, \minew{subsequent methods \cite{b7,b8,b11,b29,b59} have leveraged deep convolutional neural networks (CNNs) to estimate \emph{t}(\emph{x}) and \emph{A}. While estimating such physical parameters accurately can still be challenging due to the lack of ground truth data. In response to this challenge, recent works \cite{b62,b16,b17,b50} have shifted their focus towards end-to-end networks to directly learn the hazy-to-clear translation. Several works \cite{b19,b21,b20,b52,b55} also introduced vision transformer to improve the dehazing performance.} However, even with these advancements, in situations with dense haze, the restored images still exhibit residual fog due to the limited information provided by a single modality. \minew{Seen in Fig. 1 (a), DeHamer \cite{b21} was unable to remove dense haze and restore distant objects.}

\minew{As infrared wavelengths have a higher capability to penetrate through atmospheric particles compared to visible light, which allows infrared information to capture details that are otherwise obscured or distorted by haze in visible images, several works attempted to incorporate infrared modality to restore clean images. In their early studies, researchers \cite{b31,b30,b61} employed visible-infrared fusion technologies by transforming the color space, which pose challenges in preserving details when facing with dense haze. To tackle this issue, \cite{b33,b34} leveraged CNNs along with attention mechanisms to extract adaptive weight maps, which further enhance the fusion quality. Despite the acknowledged robust perception performance of infrared images in adverse foggy weather conditions, previous approaches have neglected to} extract deep features from the infrared images or consider complementary fusion. In short, they \minew{have primarily} treated the infrared modality as auxiliary information and did not fully leverage the advantages of each modality or explore a deep fusion of the two modalities. \minew{Motivated by these considerations, we aim to present an innovative framework for image dehazing through visible-infrared fusion. To achieve this, we commence by employing a dual-branch feature extraction network to explore deep structural features of each modality individually. Subsequently, an inconsistency fusion strategy is designed to dynamically adjust the fusion weights based on the degree of inconsistency among the features. Finally, we employ supervised learning technology to recover haze-free images, utilizing a global loss function.}

Furthermore, existing foggy datasets for deep learning-based dehazing networks are dominated by single modality, such as SOTS \cite{b26} and Foggy Cityscapes \cite{b27}. While for foggy multimodal datasets \cite{b39, b40, b41}, Bijelic \emph{et al}. \cite{b39} created the first large multimodal dataset in adverse weather for object detection. Likewise, Wang \emph{et al}. \cite{b41} constructed a visible–infrared multimodal dataset with various fog densities, and it was primarily intended for visibility range estimation. However, these datasets lack ground truth of the hazed images, as acquiring aligned image pairs under the same scene presents a significant challenge. The deficiency of  multimodal hazy/clear image pairs makes it challenging to verify the feasibility and reliability of the multimodal dehazing methods. \minew{Motivated by this, based on the AirSim simulation platform, we provide a visible-infrared dataset for image dehazing to validate the effectiveness of the proposed network.}

\begin{figure}[!t]
\centerline{\includegraphics[width=\columnwidth]{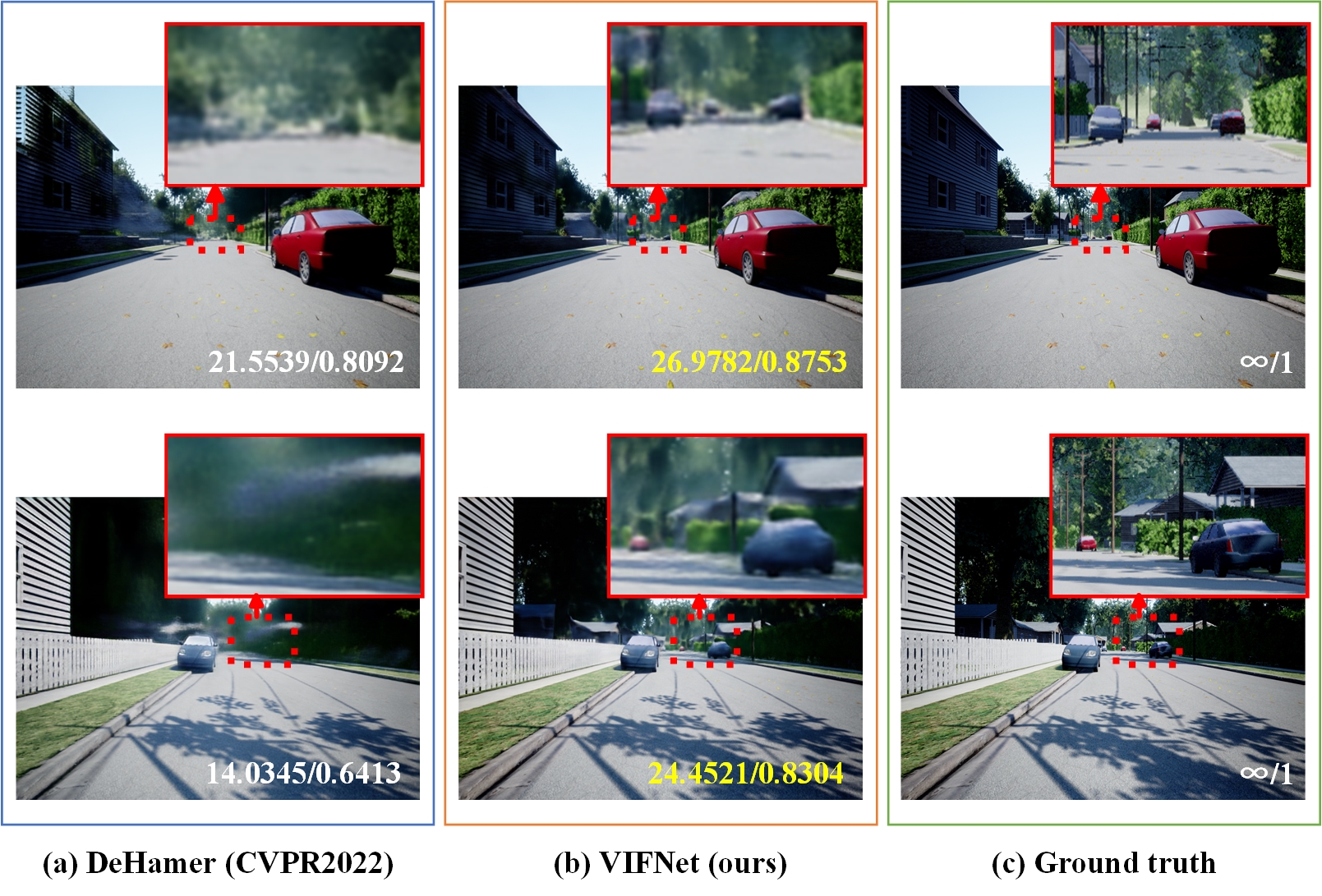}}
\caption{Comparative results of dehazing networks on \minew{the proposed AirSim-VID dataset}. The first column is the result of the single image dehazing network DeHamer \cite{b21} (SOTA), the second column is derived from the proposed VIFNet, and the last column is the ground truth. The enlarged red boxes highlight the superiority of the proposed VIFNet. }
\label{fig1}
\end{figure}

\minew{In summary, the main novelty of this paper is to design an end-to-end multimodal dehazing network that can explore deep fusion between visible and infrared modalities and make full use of the advantage of each modality.} \minew{For this paper, the main contributions are as follows:}

1) We propose an end-to-end multimodal fusion dehazing framework to restore high-quality images. \minew{In addition, we provide a visible-infrared dataset for image dehazing based on AirSim, named AirSim-VID, which contains 3 different fog concentration types.}

2) \minew{In the deep feature extraction stage, we present a Deep Structure Feature Extraction (DSFE) module, which incorporates Channel-Pixel Attention Block (CPAB) to explore more spatial and marginal information within the feature maps.}

3) \minew{In the feature weighted fusion stage, an efficient inconsistency fusion strategy is introduced to adjust the fusion weights between two modalities, which emphasizes more reliable and consistent information.}

The rest of this paper is organized as follows. Section 2 describes recent related works. Section 3 demonstrates the proposed methodology. Section 4 shows the qualitative and quantitative experiments and results on the proposed dataset. Finally, Section 5 concludes our work and remaining issues.

\section{Related works}

\subsection{Single Image Dehazing}
\minew{The primary objective of single image dehazing is to restore high-quality images in hazy conditions.} Existing image dehazing methods can be broadly categorized into handcrafted prior-based methods and deep learning-based methods.

\begin{figure*}[ht!]
\centerline{\includegraphics[width=\textwidth]{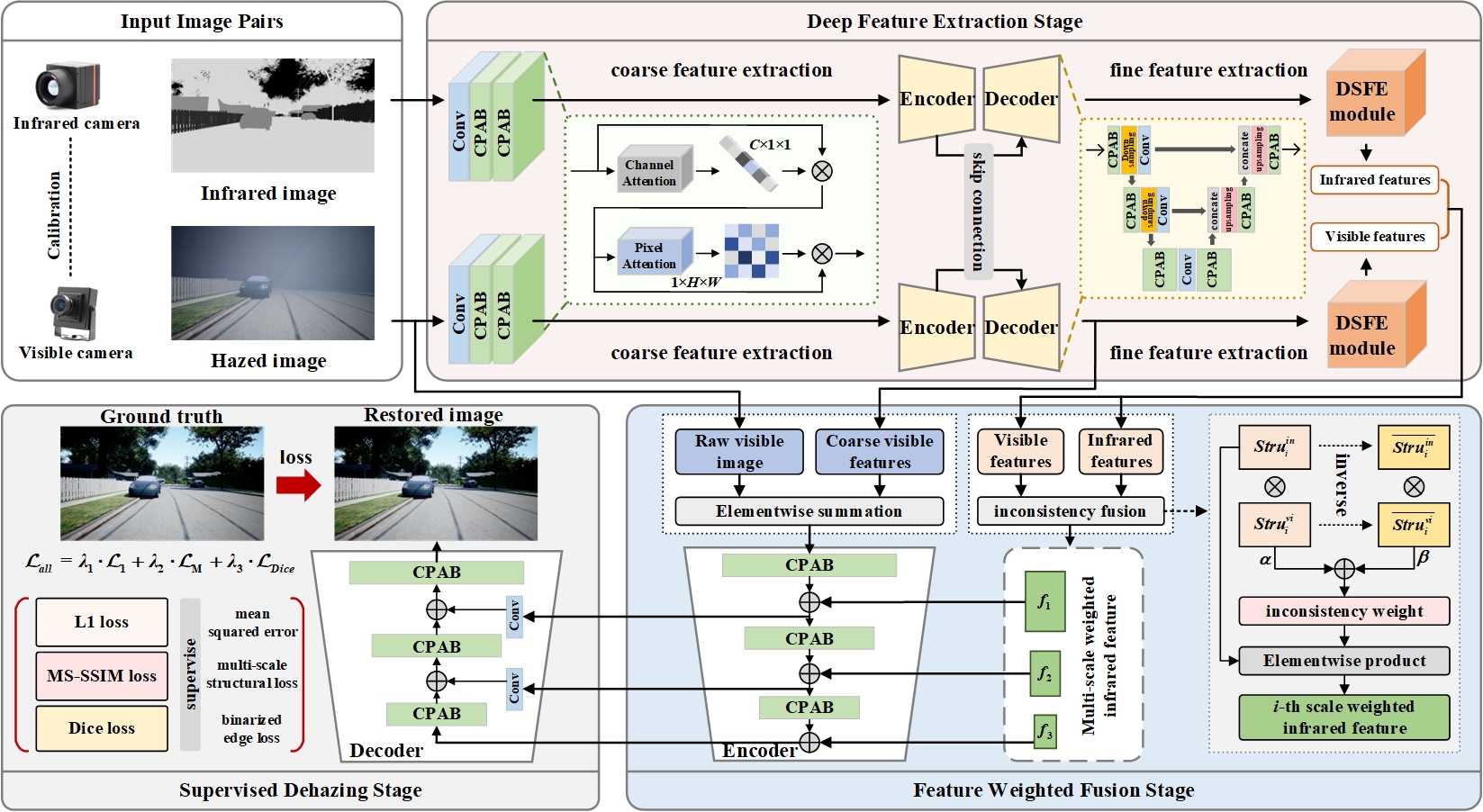}}
\caption{Overall architecture of the proposed VIFNet. In the deep feature extraction stage, an encoder-decoder architecture and DSFE module are adopted to extract multi-scale structure features from coarse to fine. Then, the multi-scale deep structure features are fused by applying the inconsistency fusion strategy and subsequently aggregated into the encoder, together with the summation of raw visible images and coarse visible features. Finally, the training process is supervised by a combined loss function.}
\label{fig2}
\end{figure*}

\subsubsection{Handcrafted Prior-based Image Dehazing Methods}
On the basis of atmospheric scattering theory \cite{b1}, these methods usually adopt handcraft priors from empirical observations. \minew{Along this line,} Tan \cite{b2} considered that images with enhanced visibility exhibit higher contrast and the atmospheric light varies smoothly across small pixel regions, then presented the dehazing method by maximizing the local contrast of the restored image. \minew{Subsequently, a variety of priors are proposed. He \emph{et al}. \cite{b3} proposed dark channel prior (DCP) with the assumption that the pixels in non-haze regions have low intensity in at least one color channel.} Zhu \emph{et al}. \cite{b4} developed a scene depth estimation model for haze removal with color attenuation prior. Berman \emph{et al}. \cite{b5} utilized non-local prior to recover clean images, assuming that the colors of a haze-free image can be approximated by distinct colors in RGB space. Additionally, haze-lines prior \cite{b6} was introduced to estimate airlight. However, the priors heavily rely on assumptions that are scene-specific. For instance, dark channel prior \cite{b3} incongruously treats sky regions, resulting in large areas of texture and fragmentation after haze removal. 

\subsubsection{Deep learning-based Image Dehazing Methods}
\minew{With the swift advancement of CNNs, propelled by the availability of extensive datasets, these algorithms harness} deep CNNs for two distinct purposes. One focuses on estimating the key parameters \minew{(i.e., transmission map \emph{t}(\emph{x}) and global atmospheric light \emph{A})} of the atmospheric scattering model, while the other directly learns the translation between hazy and clear images. 

For the former, Cai \emph{et al}. \cite{b7} \minew{firstly proposed DehazeNet, a trainable CNN for medium transmission map estimation that is subsequently used to recover the haze-free image via an atmospheric scattering model.} Similarly, Ren \emph{et al}. \cite{b8} presented a multi-scale convolutional neural network (MSCNN) for coarse-to-fine regression of the transmission maps. Lately, researchers \cite{b10,b11,b12} adopted updater networks to smooth out the transmission map or atmospheric light using iterative optimization methods. \minew{More recently, several studies \cite{b28,b29,b59}} exploited different generators to estimate the physical parameters separately. However, \minew{it is hard to accurately estimate such physical parameters, as Obtaining ground truth data for these parameters is difficult in real-world scenarios, posing challenges in training and validating models effectively.} \minew{Therefore, for the latter approaches, there is an emphasis on utilizing end-to-end models. AODNet \cite{b62} was the pioneering method that employed a lightweight CNN to directly generate a clean image. Building upon this, subsequent advancements} introduced adaptive feature fusion attention modules to enhance the flexibility of the networks. \minew{For instance, FFANet \cite{b16} incorporated feature attention module to adaptively highlight critical features by assigning varying weight coefficients to each channel and pixel. Similarly, AECRNet \cite{b17} further employed contrastive regularization as opposing forces, and USIDNet \cite{b50} conducted disentangled representations through a compact multi-scale feature attention module. In recent years,} Vision Transformer (ViT) has been introduced to improve dehazing performance \minew{\cite{b19,b21,b20,b52,b55}. Dehazeformer \cite{b19} modified the Swin Transformer by considering aspects such as normalization layer, activation function, and spatial information aggregation scheme. DeHamer \cite{b21} embedded prior haze density into the position encoder, further enhancing the dehazing process. Additionally, researchers have explored using domain adaptation techniques \cite{b49,b56} to improve the generalization of deep learning-based dehazing models, aiming to enhance their performance on real-world hazy images. However, even with the advancements made in deep learning-based dehazing methods, restoring images under dense haze conditions remains challenging due to the limited information provided by a single modality. As a result, the restored images may still exhibit residual fog or haze artifacts, affecting the overall quality of the dehazed output.} 

\subsection{\minew{Visible-infrared Fusion for Image Dehazing}}
\minew{Infrared light possesses superior penetration ability compared to visible light, leading to higher contrast and sharper edges in hazy conditions. Consequently, fusing infrared information for color image dehazing emerges as a promising approach. In the initial stages, simple fusion technique, such as Bayes’ theorem \cite{b31}, was employed to combine the information from both modalities. Subsequent studies introduced more sophisticated fusion algorithms through high-frequency components analysis \cite{b30}, Laplacian–Gaussian pyramid method \cite{b32}, and color regularization \cite{b61}. These methods aimed to leverage the complementary characteristics of visible and infrared images, such as their different luminance to haze and scene details, to achieve enhanced dehazing results. Nonetheless, due to their primary focus on image processing in the HSV color space or RGB color space, these methods had limitations in preserving details of distant objects and reducing color distortion, particularly when dealing with scenes with higher fog concentrations.} 

\minew{Furthermore, with the advent of deep learning approaches, convolutional neural networks (CNNs) have been applied to learn the optimal fusion weights. For example,} Qin \emph{et al}. \cite{b33} designed multiple CNN dehazing units to extract adaptive weight maps that capture the haze distribution. Via the channel-attention structure and residual learning model, \minew{Guo \emph{et al}. \cite{b34} presented an end-to-end RSDehazeNet for haze removal. Similarly, Ma \emph{et al}. \cite{b35} constructed multiple branches and employed different attention modules to transfer the useful information among the spectral bands. Moreover, Xie \emph{et al}. \cite{b57} utilized the regional contrast information of the infrared image to guide the contrast enhancement and transmission map refinement. Despite the advancements achieved in visible-infrared image fusion for dehazing, there still exist challenges to be addressed, such as the handling of preserving fine image details. This limitation may arise from the fact that existing methods primarily treat infrared images as guided information, without fully leveraging the rich and detailed multi-scale features offered by infrared data or adequately addressing the inherent inconsistencies between the two modalities.}  

In contrast to the aforementioned approaches, we propose an end-to-end multimodal fusion network for image dehazing by exploring structural differences between visible and infrared images. Furthermore, we fuse multi-scale deep structural features from both modalities using inconsistency weights to preserve valuable information effectively.

\section{Proposed Method}
In this section, the overall framework of visible-infrared fusion network (VIFNet) for image dehazing is proposed. Besides, the design of DSFE module is deduced and the inconsistentcy fusion strategy is presented.

\subsection{\minew{Overview of VIFNet}}
The overall architecture of the proposed VIFNet is illustrated in Fig. 2 and consists of three stages: deep feature extraction stage, feature weighted fusion stage, and supervised dehazing stage. 

\begin{algorithm}[t]
	\renewcommand{\algorithmicrequire}{\minew{\textbf{Input:}}}
	\renewcommand{\algorithmicensure}{\minew{\textbf{Output:}}}
	\caption{\minew{Pseudocode of Visible-Infrared Fusion Network for Image Dehazing}}
	\label{algorithm: masp}
	\begin{algorithmic}[1]
		\REQUIRE
			\minew{Visible-infrared image pairs $\mathbf{P}$=($\mathbf{I^{vi}}$, $\mathbf{I^{in}}$),
                ground truth $\mathbf{Y}$, initial network parameters $\Theta$, total loss ${L}_{all}$, initial learning rate $\eta$, training epochs $\mathbf{N_{epochs}}$.}
            
		\ENSURE
			\minew{predicted dehazed image $\mathbf{X}$, trained network parameters $\hat{\Theta}$.}
        \color{black}
        \REPEAT
        \FOR {$n=1$ to $\mathbf{N_{epochs}}$}
        \STATE // Stage 1. Calculate deep structure features.
        \STATE $F_{ED}^{vi}, F_{ED}^{in}\leftarrow \mathbf{Encoder\_ Decoder}(\mathbf{I^{vi}}, \mathbf{I^{in}})$;
        \STATE $Stru_i^{vi}, Stru_i^{in}\leftarrow \mathbf{DSFE}(F_{ED}^{vi}, F_{ED}^{in})$;
        \STATE // Stage 2. Calculate weighted features through inconsistency fusion stragety.
        \STATE $f_i \leftarrow {\cal F}(Stru_i^{vi}, Stru_i^{in})\otimes Stru_i^{in}$; 
        \STATE $W_i \leftarrow \mathbf{Encoder2}(f_i, F_{ED}^{vi}\oplus \mathbf{I^{vi}})$;
        \STATE // Stage 3. Supervised training.
        \STATE $\mathbf{X}\leftarrow \mathbf{Decoder2}(W_i)$;
        \STATE ${loss}(\Theta)\leftarrow {L}_{all}(\mathbf{X}, \mathbf{Y})$;
        \STATE $\Theta \leftarrow \Theta - \eta \nabla{loss}(\Theta)$
        \ENDFOR
        \UNTIL {converged}
        \RETURN $\hat{\Theta}\leftarrow \Theta$
	\end{algorithmic}
\end{algorithm}

The first deep feature extraction stage serves as the basis of the dehazing process, aiming to extract more discriminative structure features from visible and infrared images. \minew{Here, we utilise a dual-branch architecture to independently extract features from the visible and infrared images. Following the coarse-to-fine feature extraction process, each branch employs} an encoder-decoder architecture to obtain coarse features, where skip connection is used to introduce shallow convolution layer features into the upsampling or deconvolution process, \minew{thus acquiring} multi-scale and multi-level information with high spatial resolution. \minew{Furthermore, we incorporate the} Channel-Pixel Attention Block (CPAB) to enhance the capture of edges, textures, and dense hazy areas. \minew{Then Deep Structure Feature Extraction (DSFE) module is designed to extract fine features from both modalities.}

The second feature weighted fusion stage intends to combine the extracted features with different weights according to their advantageous information, which involves two steps. In the initial step, we fuse the multi-scale deep structure features from visible and infrared modalities using an inconsistency fusion strategy. \minew{This  strategy calculates the weight map of the infrared structure features at each scale. In the second step, we take the elementwise summation of original visible image and coarse visible features as the input for the encoder. Subsequently, we fuse the weighted infrared features at each scale. As a result, this stage generates multi-scale multimodal fusion features.}

The last supervised dehazing stage utilizes a decoder to restore haze-free image while being supervised with a global loss function. At each upsampling stage, the encoded multi-scale fusion features are skip-connected with the corresponding decoded features. To accelerate convergence and minimize loss during trainin, we combine multiple loss functions, including L1 loss ($\mathcal{L}_1$), MS-SSIM loss ($\mathcal{L}_{\mathrm{M}}$), and Dice loss ($\mathcal{L}_\textit{Dice}$), using different coefficients. \minew{These loss functions effectively preserve multi-scale structural information and binarized edge information during the dehazing process. Algorithm 1 provides a pseudocode outline for VIFNet.}

\subsection{DSFE Module}

\minew{To extract fine features from the upper coarse feature extraction stream, we introduce a novel Deep Structure Feature Extraction (DSFE) module. This module is designed to enhance the association of contextual and multi-scale spatial characteristics. It leverages features from both the encoder and decoder to extract deep structure features, thereby capturing more perceptual information. Fig. 3 exhibits the detailed module frame, illustrating the components and their connections. The entire process can be summarized as follows.}

\begin{figure}[t]
\centerline{\includegraphics[width=0.80\columnwidth]{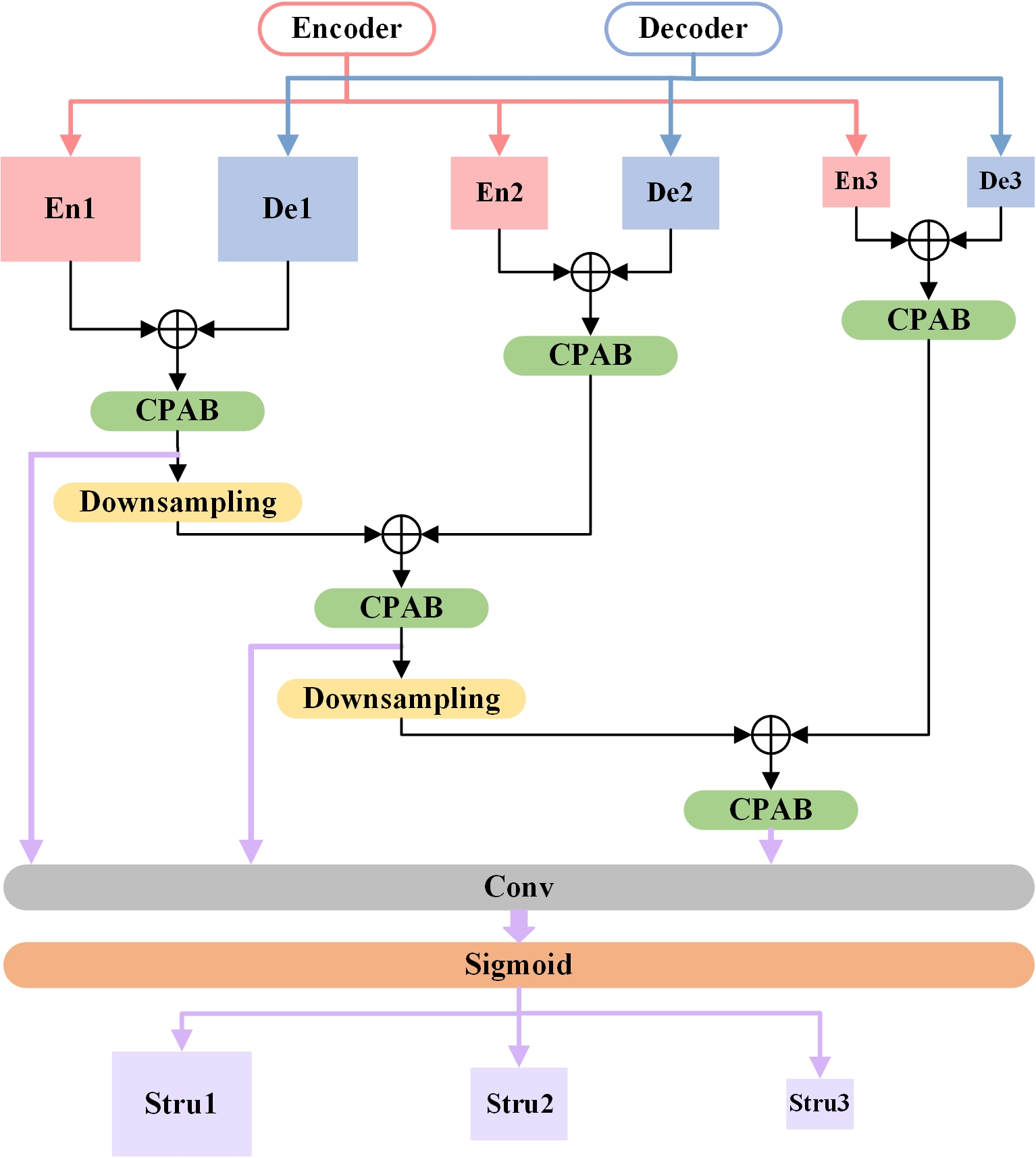}}
\caption{Detailed frame of the Deep Structure Feature Extraction (DSFE) module. The multi-scale encoded and decoded feature maps are regarded as input, and the module outputs the deep structure feature maps of three different scales.}
\label{fig3}
\end{figure}

To begin with, \minew{the encoded and decoded feature maps of the \textit{i}-th scale,} denoted as $F_{Eni}$ and $F_{Dei}$, are input into the DSFE module. By concatenating feature maps of the same scale, more complete contextual information is connected. The concatenated feature map $F_{EDi}$ of the \textit{i}-th scale can be calculated as: 
\begin{equation}{{F_{EDi}} = {F_{Eni}} \oplus {F_{Dei}}}(i = 1,2,3).\label{eq3}\end{equation}

Then, CPAB is used to adjust the weights for each channel. \minew{Unlike Feature Attention (FA) module used in FFANet \cite{b16}, we replace the ReLU activation function with the PReLU activation function,} which can adaptively learns the parameters of the correction linear units and improves the accuracy with negligible additional computational costs. Mathematically, it can be expressed as:
\begin{equation}
P(k_j)=
\begin{cases}
k_j & k_j \textgreater 0\\
{a_j}{k_j} & k_j \leq 0
\end{cases}
.\label{eq4}
\end{equation}
where \minew{$k_j$} is the input of \textit{j}-th channel, $a_j$ is the negative slope of the activation function. For each channel, there is a learnable parameter to adjust the slope.

Considering the interaction between features at different scales, the upper-level feature maps obtained after downsampling is then concatenated with the lower-level feature maps. This allows for the preservation of high spatial resolution information. Afterwards, CPAB is applied again to rescale the features by considering interdependencies among feature channels, which helps in adjusting the weights of each channel to optimize the feature representation. \minew{Finally, the deep structure features of three scales are obtained through a convolution layer followed by a Sigmoid activation function layer.} Concretely, the calculation of the \textit{i}-th scale deep structure feature $Stru_i (i=1, 2, 3)$ can be noted as:
\begin{equation}\left\{\begin{array}{l}
{Stru_1}{\rm{ = }}\sigma {\rm{(\mathbb{B}}}({F_{ED1}}){\rm{)}}\\
{Stru_2}{\rm{ = }}\sigma {\rm{(\mathbb{B}}}({\rm{\mathbb{B}}}( \downarrow ({F_{ED1}}) \oplus {F_{ED2}})){\rm{)}}\\
{Stru_3}{\rm{ = }}\sigma {\rm{(\mathbb{B}}}({\rm{\mathbb{B}}}( \downarrow ({\rm{\mathbb{B}}}( \downarrow ({F_{ED1}}) \oplus {F_{ED2}})) \oplus {F_{ED3}})){\rm{)}}
\end{array}\right.\label{eq5}\end{equation}
where $\sigma$ denotes the combination of the final convolution layer and Sigmoid function layer, and $\mathbb{B}$ represents the CPAB. Besides, the downsampling process is symbolized as $\downarrow$.

\begin{figure}[t]
\centerline{\includegraphics[width=\columnwidth]{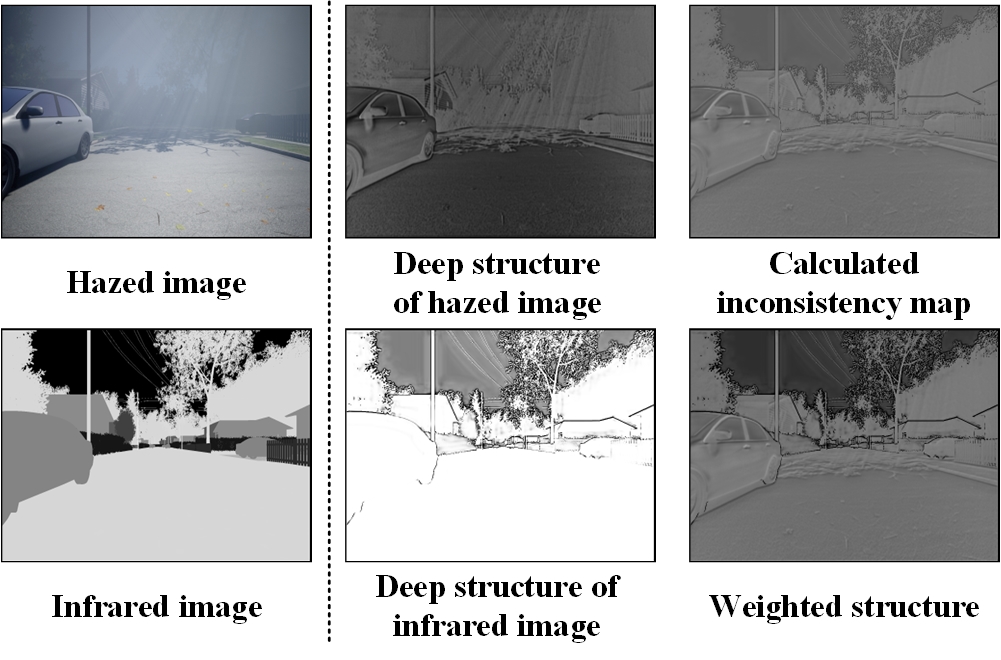}}
\caption{Visualization of deep structure feature maps of the hazed visible and infrared images, the calculated inconsistency feature map, and weighted structure feature map. \minew{With inconsistency fusion strategy, the weighted feature map enhances the overall structural information.}}
\label{fig4}
\end{figure}

\subsection{Inconsistency Fusion Strategy}

To address the issue of haze blur in visible images, as well as the poor resolution and contrast in infrared images, a fusion strategy is proposed to integrate the complementary characteristics of these two modalities. The fusion strategy consists of two steps. Originally, pixel multiplication is used to capture the contrast difference between the two images. Build upon previous work \cite{b36}, we design an inconsistency function ${{\cal F}_l}( \cdot , \cdot )$ to calculate the inconsistency weight. This weight reflects the degree of inconsistency or difference between the visible and infrared images. \minew{With multi-scale deep structure features of the visible and infrared modalities, which can be denoted as $Str{u_{i}^{vi}}$ and $Str{u_{i}^{in}}$,} the inconsistency structure feature of the \textit{i}-th scale can be computed as:
\begin{equation}
{{\cal F}_i}\left( {Str{u_{i}^{vi}},Str{u_{i}^{in}}} \right)  =  \alpha Str{u_{i}^{vi}} \cdot Str{u_{i}^{in}}  + \beta \overline{Str{u_{i}^{vi}}} \cdot \overline{Str{u_{i}^{in}}}.\label{eq6}  
\end{equation}
where $\alpha$, $\beta$ represent the corresponding weight of each items. \minew{Here, $\overline{Stru_{i}^{vi}}$ and $\overline{Stru_{i}^{in}}$ mean the inverse operation of the deep structure feature for visible and infrared modalities,} with the purpose of making full use of redundant complementary information contained in inverse images. 

Then, we utilize elementwise product for the \textit{i}-th scale feature between inconsistency feature ${\cal F}_i$ and deep structure of the infrared image $Stru_{i}^{in}$, and the weighted structure feature $f_i$ is obtained, as described below.
\begin{equation}{f_i} = {{\cal F}_i} \otimes Stru_i^{in}.\label{eq7}\end{equation}

To visually illustrate the effectiveness of the DSFE module, we print the deep structure feature maps of the visible image and infrared image, as well as the calculated inconsistency feature map and weighted structure feature map in Fig. 4. 
\minew{It is evident that the deep structure feature map of the visible image capture rich and detailed edges and textures of the objects from close range, a distance that is not obscured by the dense haze. On the other hand, the deep structure feature map of the infrared image represents the areas that are located farther away from the viewpoint. The inconsistency map effectively highlights the disparities between the two modalities, revealing the locations where they diverge in terms of structural information. By applying the inconsistency fusion strategy, the weighted structure feature combines the strengths of each modality, thereby enhancing the overall structural feature. As a result, the outlines of the objects become more distinct, and the contrast between the sky areas and other regions becomes more prominent, resulting in a noticeable visual distinction.}

\begin{figure*}[htb]
\centerline{\includegraphics[width=\textwidth]{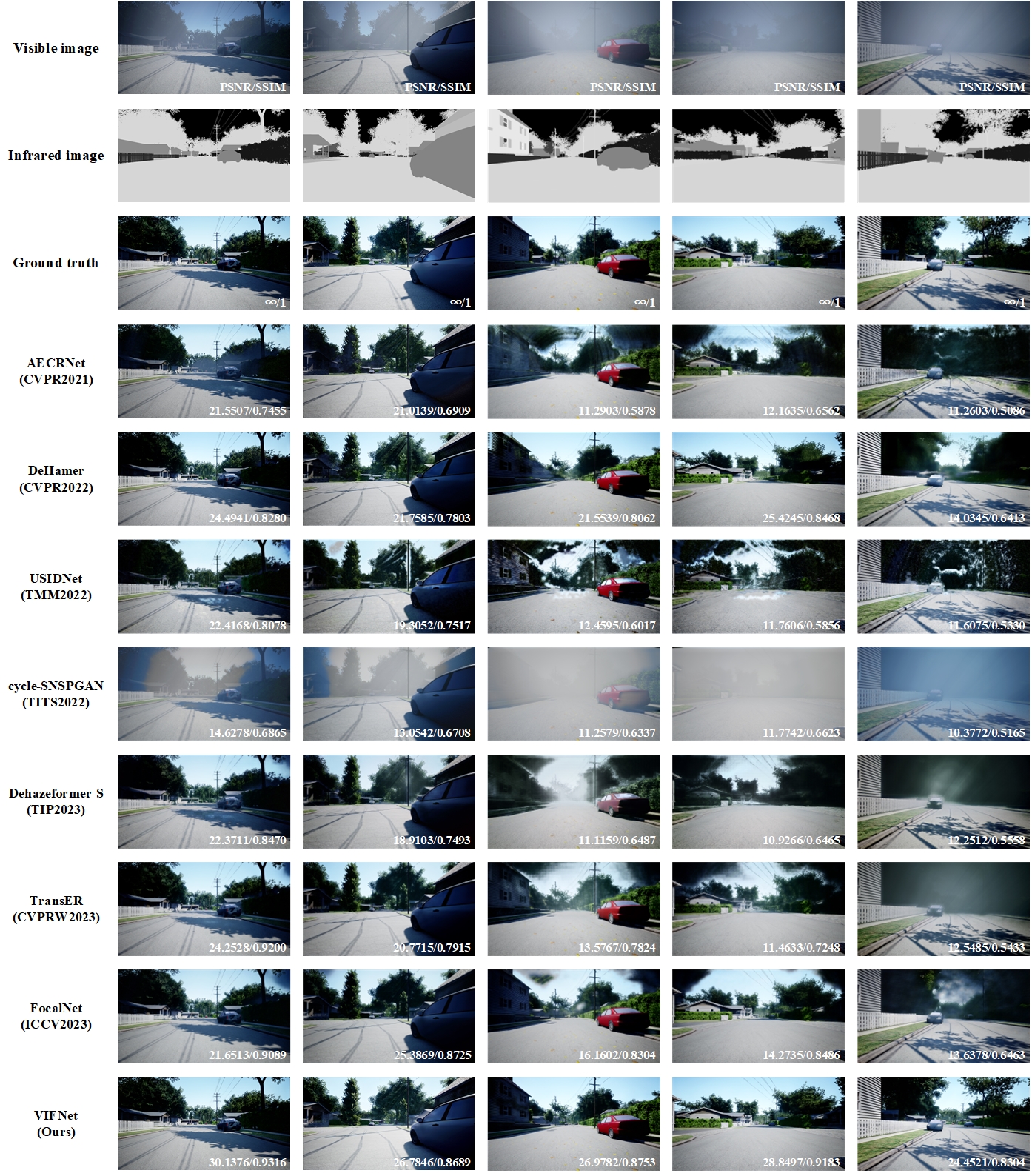}}
\caption{\minew{Comparison of dehazing results on the AirSim-VID dataset. The first two columns, the middle two columns, and the last column represent mist, medium haze, and dense haze, respectively.}}
\label{fig5}
\end{figure*}

\subsection{Loss Function}

Mean squared error (MSE), namely L1 loss, is the most widely used loss function for image dehazing tasks. Given the ground-truth \textit{Y} and the predicted image \textit{X}, L1 loss ($\mathcal{L}_1$) can be expressed as:
\begin{equation}\mathcal{L}_1=\dfrac{1}{N}\sum_{i=1}^N\Vert Y_i-X_i \Vert.\label{eq8}\end{equation}

To further enhance the boundary of multi-layer structures, we apply the multi-scale structural similarity index (MS-SSIM) \cite{b37} loss function to assign higher weights to the fuzzy boundary, the MS-SSIM loss ($\mathcal{L}_{\mathrm{M}}$) function is defined as:
\begin{equation}\mathcal{L}_{\mathrm{M}}=1-\prod_{m=1}^{M}\left(\frac{2 \mu_{x} \mu_{y}+C_{1}}{\mu_{x}^{2}+\mu_{y}^{2}+C_{1}}\right)^{\beta_{m}}\left(\frac{2 \sigma_{x} \sigma_{y}+C_{2}}{\sigma_{x}^{2}+\sigma_{y}^{2}+C_{2}}\right)^{\gamma_{m}}.\label{eq9}\end{equation}
where \textit{M} represents the total number of the scales, $\mu_x$, $\mu_y$ and $\sigma_x$, $\sigma_y$ are the mean and standard deviations of \textit{x} and \textit{y}, respectively, and $\sigma_{xy}$ denotes their covariance. The parameters $\beta_m$ and $\gamma_m$ mean the relative importance of the two components in each scale. Beyond that, $C_1$ and $C_2$ are two small constants to avoid the unstable circumstance of dividing by zero.

We also introduce Dice loss as the training loss to enhance the supervision of fuzzy boundaries, which is proposed by Deng \textit{et al}. \cite{b38}. The total Dice loss is described as:
\begin{equation}\mathcal{L}_{Dice}=\sum_{i=1}^{3}\mathrm{Dice}(edge_{i}^{out},edge_{i}^{gt}).\label{eq10}\end{equation}
where \textit{i} represents the \textit{i}-th channel of the image, $edge_{i}^{out}$ and $edge_{i}^{gt}$ stand for the binarized edge maps of the predicted image and the ground-truth image, which are obtained by Sobel operator. For each channel, the Dice loss is calculated by:
\begin{equation}{\rm{Dice(}}edg{e^{out}}{\rm{,}}edg{e^{gt}}{\rm{) = }}\frac{\sum\limits_{j=1}^N {{\left( {edge_j^{out}} \right)}^2} + \sum\limits_{j=1}^N {{\left( {edge_j^{gt}} \right)}^2} + {C_3}}{{2 \times \sum\limits_{j=1}^N {edge_j^{out}edge_j^{gt}}+{C_3}}}.\label{eq11}\end{equation}
where $edge_{j}^{out}$ and $edge_{j}^{gt}$ are the \textit{j}-th pixel on the predicted image and the ground-truth image, and $C_3$ is added to avoid zero probability on the basis of Laplacian smoothing.

By combining L1 loss ($\mathcal{L}_1$), MS-SSIM loss ($\mathcal{L}_{\mathrm{M}}$), and edge loss ($\mathcal{L}_\textit{Dice}$), we develop a pixel-scale-structure level hybrid loss for visible-infrared image dehazing, which is capable of capturing both multi-scale and fine structures with clear boundaries. Then, the total loss function ($\mathcal{L}_{all}$) in the training phase is formulated as:
\begin{equation}\mathcal{L}_{all}=\lambda_1\cdot\mathcal{L}_1 + \lambda_2\cdot\mathcal{L}_{\mathrm{M}} + \lambda_3\cdot\mathcal{L}_{Dice}.\label{eq12}\end{equation}
where $\lambda_1$, $\lambda_2$, $\lambda_3$ are the corresponding coefficients.

\section{Experiments}

\subsection{Dataset}
In this study, we conduct training and evaluation of our model on both simulated and real-world datasets to assess its performance.

\textbf{AirSim-VID.} In terms of simulated datasets, we propose a foggy visible-infrared dataset based on AirSim \cite{b42}, a high-fidelity simulation platform for autonomous vehicles, which can provide real-time ground truth and paired images under different degrees of fog conditions. The pipeline of the dataset generation is as follows. Firstly, we use the official scenario—AirSimNH (small urban neighborhood block) as the simulation scene to collect data. Next, a visible camera and an infrared camera are both mounted in the same location on the front side of an unmanned vehicle. The images are captured at intervals of 5 meters during the vehicle's movement within the specified mileage. Overall, our dataset comprises 2,310 aligned hazy/clear/infrared image pairs, each corresponding to three different fog concentration coefficients.

\textbf{\minew{NTIRE Challenge Dataset.}} \minew{Dense-Haze \cite{b43} and NH-HAZE \cite{b44} were introduced with the NTIRE 2019 and NTIRE 2020 Dehazing Challenge, respectively.} These datasets show different haze densities according to local image areas, which can reflect the ability of the model to cope with different fog concentrations. Due to the lack of infrared modalities, we use the pre-trained rgb-to-nir generative model \cite{b45} to generate infrared images.

\minew{\textbf{Natural hazy dataset.} The M3FD dataset \cite{b60} consists of 4500 registered visible-infrared image pairs captured in various real-world scenes. These image pairs are categorized into four typical types: daytime, overcast, night, and challenge. For our evaluation, we specifically focus on the challenge category, which comprises natural hazy images. This category allows us to assess the effectiveness of our method in handling challenging atmospheric conditions and improving visibility in hazy scenes.}

\subsection{Implementation Details}
All experiments were conducted by Torch 2.0.0 and Torchvision 0.15.1 with an NVIDIA RTX 3090 Ti GPU on a personal laptop. In the training process, the initial learning rate, the batch size, the training iterations, and the weight decay were set to 0.0001, 8, 100000, and $5^{-4}$, respectively. All training samples were resized to 240$\times$240. Besides, the Adam optimizer was applied with exponential decay rates $\beta_1$ and $\beta_2$ equal to 0.9 and 0.999, respectively. Moreover, cosine annealing strategy was utilized to adjust the learning rate. 

\subsection{Quantitative and Qualitative Results}

\subsubsection{Evaluation metrics}
In order to demonstrate the effectiveness of the proposed method on the above datasets, Peak Signal to Noise Ratio (PSNR) and the Structural Similarity index (SSIM) are adopted as quantitative evaluation metrics, which are commonly used to assess image quality in the context of haze removal tasks. Both PSNR and SSIM are calculated by comparing the processed image with the clean original image, which serves as a reference. It is worth noting that a higher value of PSNR and a value of SSIM closer to 1 indicate superior haze removal performance.

\begin{table*}[t]
\caption{\minew{Quantitative comparison (average PSNR/SSIM) of the dehazing results on AirSim-VID Dataset. \textbf{Bold} fonts indicate best performance, and results with \underline{underline} represent the second best. ”-” indicates no training code provided.}}
\setlength{\tabcolsep}{4pt}
\renewcommand{\arraystretch}{1.2}
\label{table}
\centering
\begin{tabular}{ccccccccccc}  
\toprule
\multirow{2}{*}{Methods} & \multirow{2}{*}{Reference} & \multicolumn{2}{c}{mist} & \multicolumn{2}{c}{medium haze} & \multicolumn{2}{c}{dense haze} & \multirow{2}{*}{\minew{\#Params(M)}} & \multirow{2}{*}{\minew{FLOPs(G)}} & \multirow{2}{*}{\minew{time(s)}} \\
            &                   & PSNR  & SSIM            & PSNR  & SSIM          & PSNR  & SSIM &  &  &       \\ 
\midrule
FFANet \cite{b16}         & AAAI’20             & 21.36 & 0.8031          & 13.26 & 0.5936        & 11.39 & 0.4764   & \minew{4.46}  &  \minew{143.9}    &  \minew{0.056}                   \\
AECRNet \cite{b17}        & CVPR’21             & \minew{\underline{25.49}} & 0.8828          & \minew{\underline{22.52}} & 0.796         & 14.44 & \minew{\underline{0.6175}}   & \minew{2.61}  &  \minew{26.1}    &  \minew{0.006}                       \\
DeHamer \cite{b21}         & CVPR’22          & 25.06 & -               & 22.13 & -             & \minew{\underline{15.66}} & -     &   \minew{29.44}   &  \minew{48.93}    &  \minew{5.29}                        \\         
\minew{USIDNet \cite{b50}}        & \minew{TMM’22}             & \minew{21.72} & \minew{0.7948}          & \minew{14.38} & \minew{0.6001}         & \minew{11.72} & \minew{0.4834}    & \minew{3.77} &  \minew{35.53}    &  \minew{0.014}                       \\
\minew{cycle-SNSPGAN \cite{b49}}        & \minew{TITS’22}             & \minew{13.40 }& \minew{0.6733}          & \minew{10.59} & \minew{0.5547}         & \minew{10.30} &  \minew{0.5382}     & \minew{2.36}  &  \minew{59.01}    &  \minew{0.190}                    \\
\minew{Dehazeformer-S \cite{b19} }       & \minew{TIP’23}     & \minew{21.85} & \minew{0.8627}         & \minew{13.42} & \minew{0.6284}         & \minew{10.74} & \minew{0.5189}        &  \minew{1.28}  &  \minew{6.565}    &  \minew{0.013}                 \\
\minew{TransER \cite{b20}}        & \minew{CVPRW’23}             & \minew{24.07}& \minew{\underline{0.9101}}          & \minew{15.08} & \minew{0.7319}         & \minew{11.50} & \minew{0.5080}                & \minew{2.60} &  \minew{14.81}    &  \minew{0.720}          \\
\minew{FocalNet \cite{b52}}        & \minew{ICCV’23}             & \minew{21.56} & \minew{0.9045}          & \minew{17.62} & \minew{\underline{0.8068}}         & \minew{12.33} & \minew{0.6113}          &  \minew{3.74}  &  \minew{30.63}    &  \minew{0.009}              \\
ours         & -                  & \textbf{27.73} & \textbf{0.9105}          & \textbf{25.53} & \textbf{0.8493}        & \textbf{24.32} & \textbf{0.8242}  & \minew{9.78} &  \minew{155.6}    &  \minew{0.145}                            \\
\bottomrule
\end{tabular}
\label{tab1}
\end{table*}

\subsubsection{Results on AirSim-VID Dataset}

In Table 1, \minew{we summarize the performance of our proposed method and several competitive methods (FFANet \cite{b16}, AECRNet \cite{b17}, DeHamer \cite{b21}, USIDNet \cite{b50}, cycle-SNSPGAN \cite{b49}, Dehazeformer-S \cite{b19}, TransER \cite{b20}, FocalNet \cite{b52}) in recent years on the AirSim-VID Dataset. To ensure a fair comparison, we trained these models using the same configuration. The results clearly demonstrate that our method consistently outperforms these methods,} achieving higher PSNR and SSIM scores across various fog concentrations, which indicates that our VIFNet effectively reduces distortion and preserves more image information. \minew{Specifically, when under the mist, our method achieves a noteworthy PSNR gain of 2.24 dB, while a remarkable PSNR gain of 3.01 dB with medium haze. Particularly, under dense hazy conditions, our proposed method exhibits the most substantial improvement, achieving an impressive PSNR gain of 8.65 dB.} Similarly, as the concentration of fog increases, the SSIM value also shows an upward trend, increasing from 0.0277 to 0.0533 to 0.2067.

 
Visual dehazing results comparison are also displayed in Fig. 5. \minew{It can be observed that existing methods fail to remove the dense haze and suffer from color distortion. While DeHamer \cite{b21} manages to maintain color space consistency, it fails to restore clear pixel regions or detect distinct edges for distant objects. In areas with high contrast, methods such as AECRNet \cite{b17}, USIDNet \cite{b50}, Dehazeformer \cite{b19}, TransER \cite{b20}, and FocalNet \cite{b52} do not fully restore object details, but instead generate gray mottled artifacts in the sky regions.} In contrast, our method exhibits similar patterns to the ground truth across different fog scenarios while preserving more image details, achieving a more natural and visually pleasing appearance with the aid of additional infrared information.

\begin{figure*}[t!]
\centering
\begin{minipage}{\linewidth}
\centering
\includegraphics[width=\textwidth]{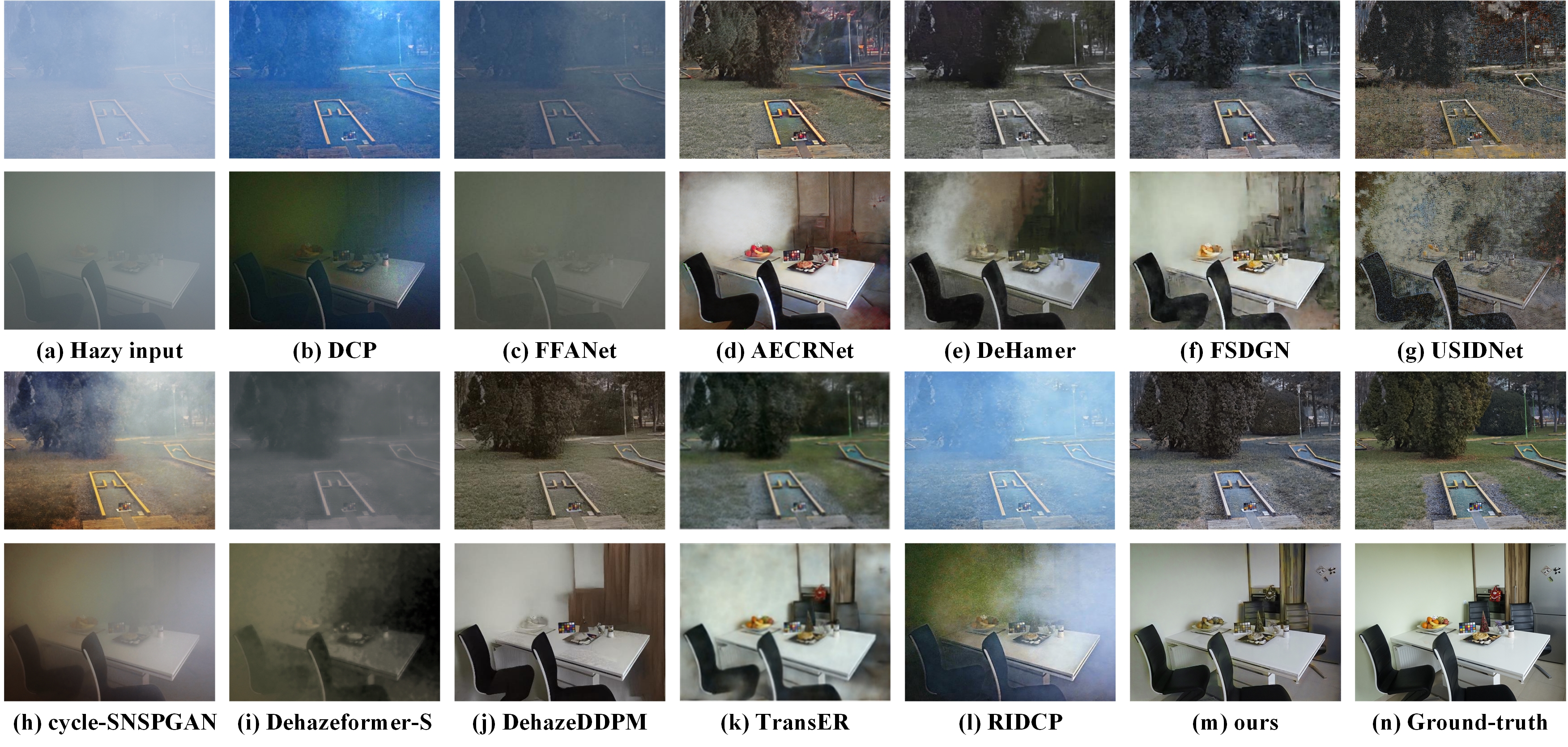}
\caption{\minew{Comparison of dehazing results on the Dense-Haze \cite{b43} dataset.}}
\label{fig7}
\end{minipage}

\vspace{1em}

\begin{minipage}{\linewidth}
\centering
\includegraphics[width=\textwidth]{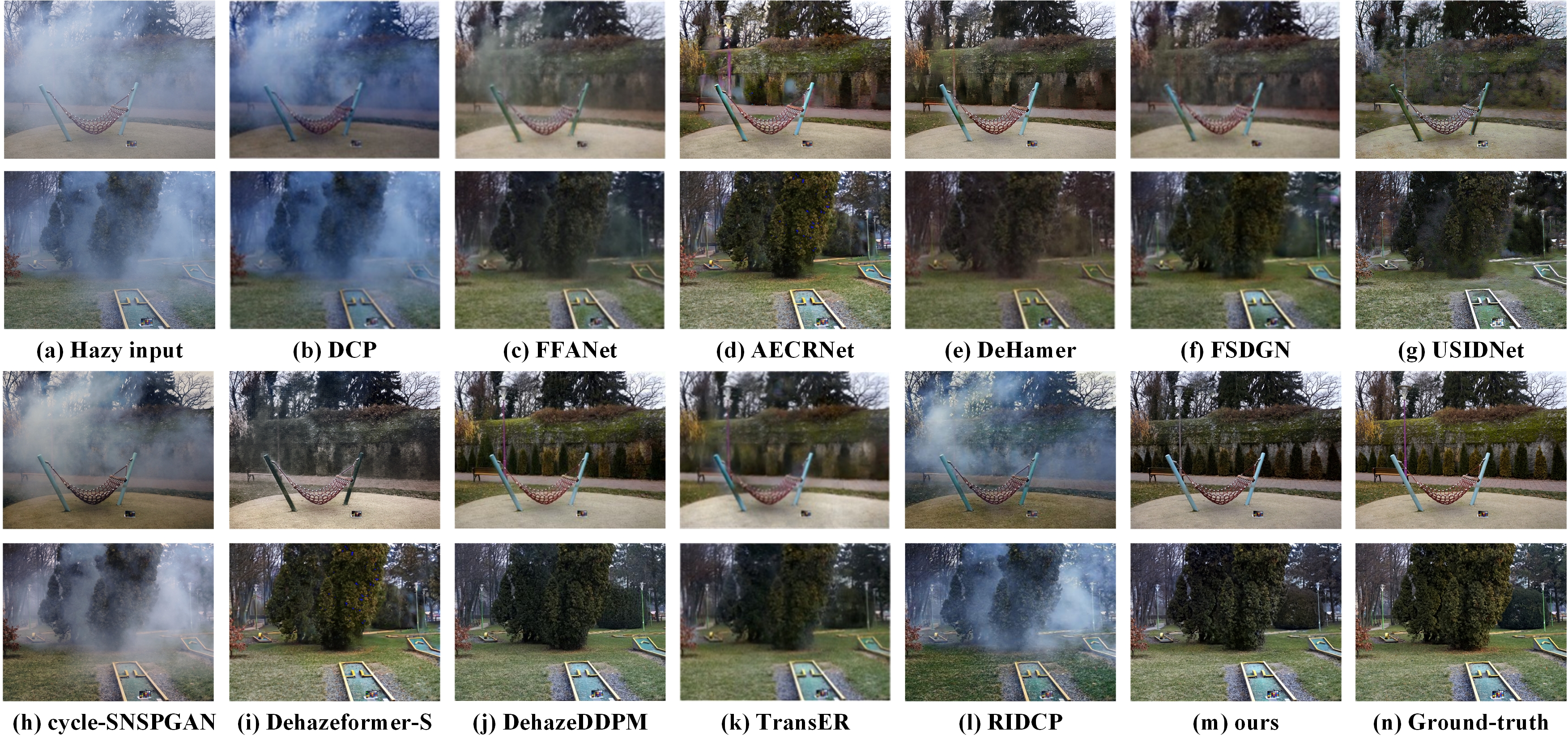}
\caption{\minew{Comparison of dehazing results on the NH-HAZE \cite{b44} dataset.}}
\label{fig8}
\end{minipage}
\end{figure*}

\subsubsection{\minew{Results on NTIRE Challenge Dataset}} 

\begin{table}[width=.9\linewidth,hb]
\caption{\minew{Quantitative comparison (average PSNR/SSIM) of the dehazing results on NTIRE Challenge Dataset. \textbf{Bold} fonts indicate best performance, and results with \underline{underline} represent the second best.}}
\label{table}
\setlength{\tabcolsep}{1pt}
\renewcommand{\arraystretch}{1.2}
\centering
\begin{tabular}{cccccc} 
\toprule
\multirow{2}{*}{Methods} & \multirow{2}{*}{Reference} & \multicolumn{2}{c}{Dense-Haze} & \multicolumn{2}{c}{NH-HAZE}   \\
              &               & PSNR  & SSIM                    & PSNR  & SSIM                                          \\ 
\midrule
DCP \cite{b3}     & TPAMI’11                    & 10.06 & 0.3856                  & 10.57 & 0.5196                                \\
FFANet \cite{b16}        & AAAI’20              & 12.22 & 0.444                   & 18.13 & 0.6473                                 \\
AECRNet \cite{b17}      & CVPR’21            & 15.80  & 0.466                   & 19.88 & 0.7073                                   \\
DeHamer \cite{b21}      & CVPR’22             & 16.62 & 0.5602                  & 20.66 & 0.6844                                  \\
FSDGN \cite{b46}       & ECCV’22              & 16.91 & 0.5806                  & 19.99 & 0.7106                                  \\
\minew{USIDNet \cite{b50}}       & \minew{TMM’22}              & \minew{16.32} & \minew{0.3686}                  & \minew{19.21} & \minew{0.5794}                                  \\
\minew{cycle-SNSPGAN \cite{b49}}       & \minew{TITS’22}              & \minew{13.01} &  \minew{0.574}                 & \minew{13.78} & \minew{0.4914}                                  \\
Dehazeformer-S \cite{b19}    & TIP’23        & 16.29 & 0.510                    & 20.47 & 0.731                                    \\
TransER \cite{b20}     & CVPRW’23                & 17.03 & \minew{0.597}                  & 21.64 & \minew{0.743}                               \\
dehazeDDPM  \cite{b47}     & arxiv’23            & \underline{19.04} & 0.5922                  & \underline{22.28} & 0.7309                               \\
RIDCP \cite{b48}     & CVPR’23               & 8.09  & 0.4173                  & 12.27 & 0.4996                                   \\ 

\minew{FocalNet\cite{b52}} & \minew{ICCV’23}               & \minew{17.07}  & \minew{\underline{0.63}}                  & \minew{20.43} & \minew{\underline{0.790}}                                 \\ 
ours          & -             & \textbf{24.62} & \textbf{0.8924}                  & \textbf{26.82} & \textbf{0.9102}             \\
\bottomrule
\end{tabular}
\label{tab2}
\end{table}

\minew{In addition to the above eight methods, we compare our method with DCP \cite{b3}, FSDGN \cite{b46}, dehazeDDPM \cite{b47}, and RIDCP \cite{b48}. Table 2} lists the results of quantitative comparison results on the real-world dataset, where our VIFNet achieves the best performance in terms of PSNR and SSIM. \minew{Specifically, on the Dense-Haze dataset, our method outperforms the second-best method by 5.58 dB in PSNR and 0.2624 in SSIM. Similarly, on the NH-HAZE dataset, our method surpasses the second-best method by 4.54 dB in PSNR and 0.1202 in SSIM.} Furthermore, Fig. 6 and Fig. 7 display the visual dehazing results on the Dense-Haze dataset and the NH-HAZE dataset, respectively. It can be observed that our VIFNet is closer to the ground-truth from texture details and structural features, despite \minew{a slight presence of color distortion. This phenomenon can be attributed to the fusion process, with the infrared modality being assigned higher weights. Since the infrared modality can not capture color information, there can be a trade-off between accurately preserving color consistency and effectively enhancing structural details.} 

\minew{Furthermore, it is important to note that the competitive methods typically struggle to effectively remove fog under dense haze conditions due to their reliance on a single modality. For example, based on the Dense-Haze dataset, DCP \cite{b3}, FFANet \cite{b16}, DeHamer \cite{b21}, USIDNet \cite{b50}, cycle-SNSPGAN \cite{b49}, Dehazeformer \cite{b19}, and RIDCP \cite{b48} tend to produce darker images with severe color distortion and low resolution. Besides, methods like AECRNet \cite{b17}, FSDGN \cite{b46}, and TransER \cite{b20} have shown relative effectiveness in dehazing. However, one limitation that can be observed in these methods is the presence of unsmooth areas when restoring background regions with dense haze. Similarly, while DehazeDDPM \cite{b47} is successful in removing large areas of fog, it often misestimates the original objects present in the scene. This issue may arise due to its utilization of a generative model, which may occasionally produce outputs that do not align with the desired semantics.} 

\minew{While based on the NH-HAZE dataset, apart from DCP \cite{b3}, cycle-SNSPGAN \cite{b49}, and RIDCP \cite{b48}, the remaining methods perform better in terms of haze removal. It is worth noting that the performance of these methods can vary depending on the specific dataset and hazy conditions. In contrast, with the help of infrared modality, our VIFNet demonstrates robustness, particularly in challenging scenarios with dense haze.}

\minew{\subsubsection{Results on Natural Hazy Dataset} 


We also conducted evaluations on the M3FD dataset using pretrained models, where Fig. 8 displays a comparison of the dehazing results obtained from various methods, including FFANet \cite{b16}, AECRNet \cite{b17}, DeHamer \cite{b21}, FSDGN \cite{b46}, USIDNet \cite{b50}, cycle-SNSPGAN \cite{b49}, Dehazeformer-S \cite{b19}, TransER \cite{b20}, and FocalNet \cite{b52}. While in real-world scenarios, none of these methods were able to remove haze effectively, which may further lead to missed detection for object detection task. In comparison, our method stands out in its ability to restore the structural details of objects in the scene, thanks to the compensation provided by the infrared modality. However, it is important to note that this advantage comes with a potential drawback, namely color distortion, in which the colors in the dehazed images may deviate from their original appearance.}

\begin{table}[width=.9\linewidth,ht]

\caption{\minew{Performance comparison of visible-infrared basic fusion and with the proposed DSFE module and inconsistency function on the AirSim-VID dataset. \textbf{Bold} fonts indicate best performance.}}
\label{table}
\setlength{\tabcolsep}{1pt}
\renewcommand{\arraystretch}{1.2}
\centering
\begin{tabular}{cccccc} 
\toprule
 & basic fusion & DSFE & function & \multicolumn{1}{c}{PSNR} & \multicolumn{1}{c}{SSIM}  \\ 
\midrule
\multirow{3}{*}{mist}   & \checkmark  &    &    & 27.02                    & 0.9001                    \\
                     & \checkmark  & \checkmark  &    & 27.58                    & 0.909                     \\
                     & \checkmark  & \checkmark  & \checkmark  & \textbf{27.61}                    & \textbf{0.9093}                    \\ 
\multirow{3}{*}{medium haze}   & \checkmark  &    &    & 25.14                    & 0.8403                    \\
                     & \checkmark  & \checkmark &    & 25.33                    & 0.8478                    \\
                     & \checkmark  & \checkmark  & \checkmark   & \textbf{25.50}                     & \textbf{0.8485}                    \\ 
\multirow{3}{*}{dense haze}   & \checkmark  &    &    & 23.87                    & 0.8159                    \\
                    & \checkmark  & \checkmark &    & 24.19                    & 0.8239                    \\
                    & \checkmark  & \checkmark  & \checkmark   & \textbf{24.27}                    & \textbf{0.8241}                    \\
\bottomrule
\end{tabular}
\label{tab3}
\end{table}

Overall, the results of both quantitative and qualitative comparisons clearly demonstrate the superiority of our proposed method over the competitive method in terms of dehazing performance. Our method not only enhances visibility and restores image details, but also achieves higher accuracy and fidelity according to objective evaluation metrics. \minew{However, it is necessary to acknowledge that our method introduces a trade-off in the form of color distortion.}

\minew{\subsubsection{Computational Complexity Analysis}

To comprehensively analyze the computational complexity of our method, we present the time consumption and the computational efficiency of all the methods. As depicted in the last three columns of Table 1, it is worth noting that while the training parameters of our model occupy 9.78M, making it the second largest among the compared methods, this increase in parameter size is essential to accommodate the additional information and complexity introduced by the fusion of multiple modalities. Furthermore, it represents a trade-off that allows our model to effectively leverage multimodal images and achieve superior results.}

\subsection{Ablation Study}
To validate the reasonableness of the proposed DSFE module, as well as the inconsistency function and combined loss function used in our method, a series of ablation experiments on the AirSim-VID dataset are conducted to demonstrate the effectiveness of each component. 

\subsubsection{Effect of DSFE and inconsistency function}
To demonstrate the superiority of the proposed DSFE module, we have integrated it into the visible-infrared basic fusion network. This network consists of two separate encoder-decoder branches, where multi-scale features are simply concatenated. Moreover, the network is supervised by L1 loss $\mathcal{L}_1$. \minew{Under the mist scenario, the integration of the DSFE module results in a significant performance improvement of 0.56 dB PSNR and 0.0089 SSIM, as indicated in Table 3.} Besides, when incorporating the inconsistency function, the performance further improves by \minew{0.59} dB PSNR and 0.0092 SSIM. As evident from the preceding Fig. 4, it is apparent that the DSFE module \minew{is capable of preserving and enhancing key structural elements of the visible and infrared images, including edges, contours, textures, and other salient features that contribute to the overall clarity and perceptual quality of the output. Afterwards, the inconsistency function quantifies the level of incongruity between the structural features of visible and infrared images, which further provides the fusion weights that highlight the most reliable information for the fusion process. }

\begin{figure*}[t!]
\centerline{\includegraphics[width=\textwidth]{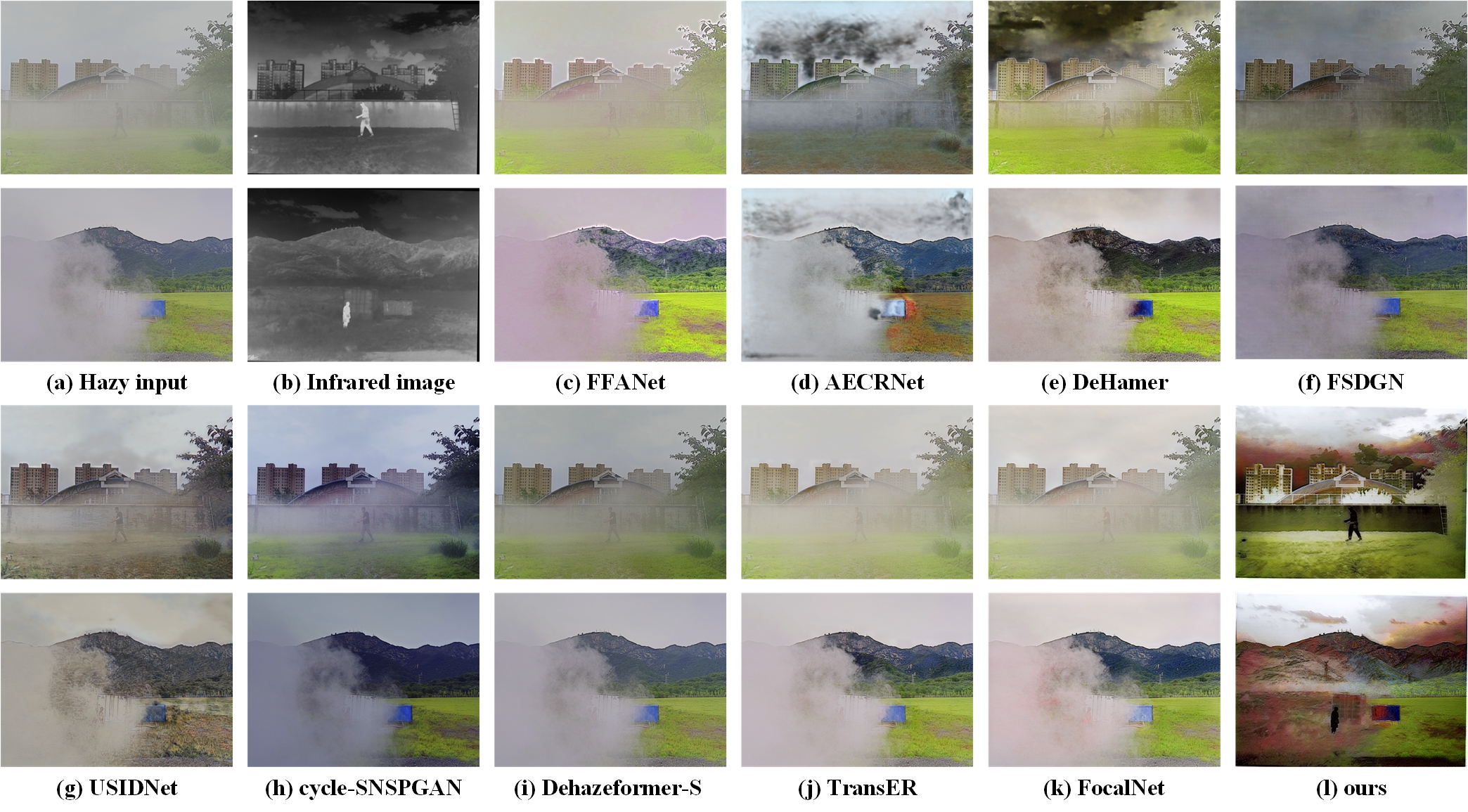}}
\caption{\minew{Comparison of dehazing results on the M3FD \cite{b60} dataset.}}
\label{fig9}
\end{figure*}

\subsubsection{Effect of $\mathcal{L}_{\mathrm{M}}$ and $\mathcal{L}_\textit{Dice}$}
To verify the effectiveness of MS-SSIM loss ($\mathcal{L}_{\mathrm{M}}$) and edge loss ($\mathcal{L}_\textit{Dice}$) during supervised training, we conduct separate training sessions on the AirSim-VID dataset using various combinations of loss functions, where $\mathcal{L}_1$ is regarded as the basic loss. \minew{The quantitative results are shown in Table 4. In terms of $\mathcal{L}_\textit{Dice}$, it focuses on preserving the edge information of objects in the image, providing valuable guidance for the restoration algorithm to recover the lost details and enhance the visibility in the visible image. As for $\mathcal{L}_{\mathrm{M}}$, by incorporating multiple scales, it captures both local and global structural similarities, providing a more comprehensive assessment of image quality.} As a result, these losses contribute to better restoration of marginal and structural details compared to the basic network, thereby enhancing the performance of image dehazing.

\begin{table}[width=.9\linewidth,!h]

\caption{\minew{Performance comparison of different loss items on the AirSim-VID dataset for $\mathcal{L}_{\mathrm{M}}$ and $\mathcal{L}_\textit{Dice}$, respectively. \textbf{Bold} fonts indicate best performance.}}
\label{table}
\centering
\renewcommand{\arraystretch}{1.2}
\begin{tabular}{cccccc}   
\toprule
                   & $\mathcal{L}_1$  & $\mathcal{L}_{\mathrm{M}}$  & $\mathcal{L}_\textit{Dice}$  & \multicolumn{1}{c}{PSNR} & \multicolumn{1}{c}{SSIM}  \\ 
\midrule
\multirow{4}{*}{\minew{mist}} &  \checkmark  &     &     & 27.61                    & 0.9093                    \\
                   &  \checkmark  &  \checkmark  &     & 27.72                    & 0.9102                    \\
                   &  \checkmark  &     &  \checkmark  & 27.68                    & 0.9100                      \\
                   &  \checkmark  &  \checkmark  &  \checkmark  & \textbf{27.73}                    & \textbf{0.9105}                    \\ 
\multirow{4}{*}{\minew{medium haze}} & \checkmark &     &     & 25.50                     & 0.8485                    \\
                   &  \checkmark  &  \checkmark  &     & 25.51                    & 0.8486                    \\
                   &  \checkmark  &     &  \checkmark & 25.52                    & 0.8489                    \\
                   &  \checkmark  &  \checkmark  &  \checkmark & \textbf{25.53}                    & \textbf{0.8493}                    \\ 
\multirow{4}{*}{\minew{dense haze}} & \checkmark &     &     & 24.27                    & 0.8241                    \\
                   &  \checkmark  &  \checkmark  &     & 24.28                    & 0.8240                     \\
                   &  \checkmark  &     &  \checkmark & 24.30                     & \textbf{0.8244}                    \\
                    &  \checkmark  &  \checkmark  &  \checkmark & \textbf{24.32}                    & 0.8242                    \\
\bottomrule
\end{tabular}
\label{tab4}
\end{table}


\minew{In addition, the ablation studies surrounding the individual components are intuitively presented in Fig. 9, providing intuitive evidence that the inclusion of these modules and losses leads to a more effective restoration of marginal and structural details compared to the basic fusion network. These findings strongly support the notion that incorporating these modules and losses significantly contributes to improved dehazing performance.}

\minew{\subsubsection{Effect of misalignment}

\begin{figure}[!ht]
\centerline{\includegraphics[width=\columnwidth]{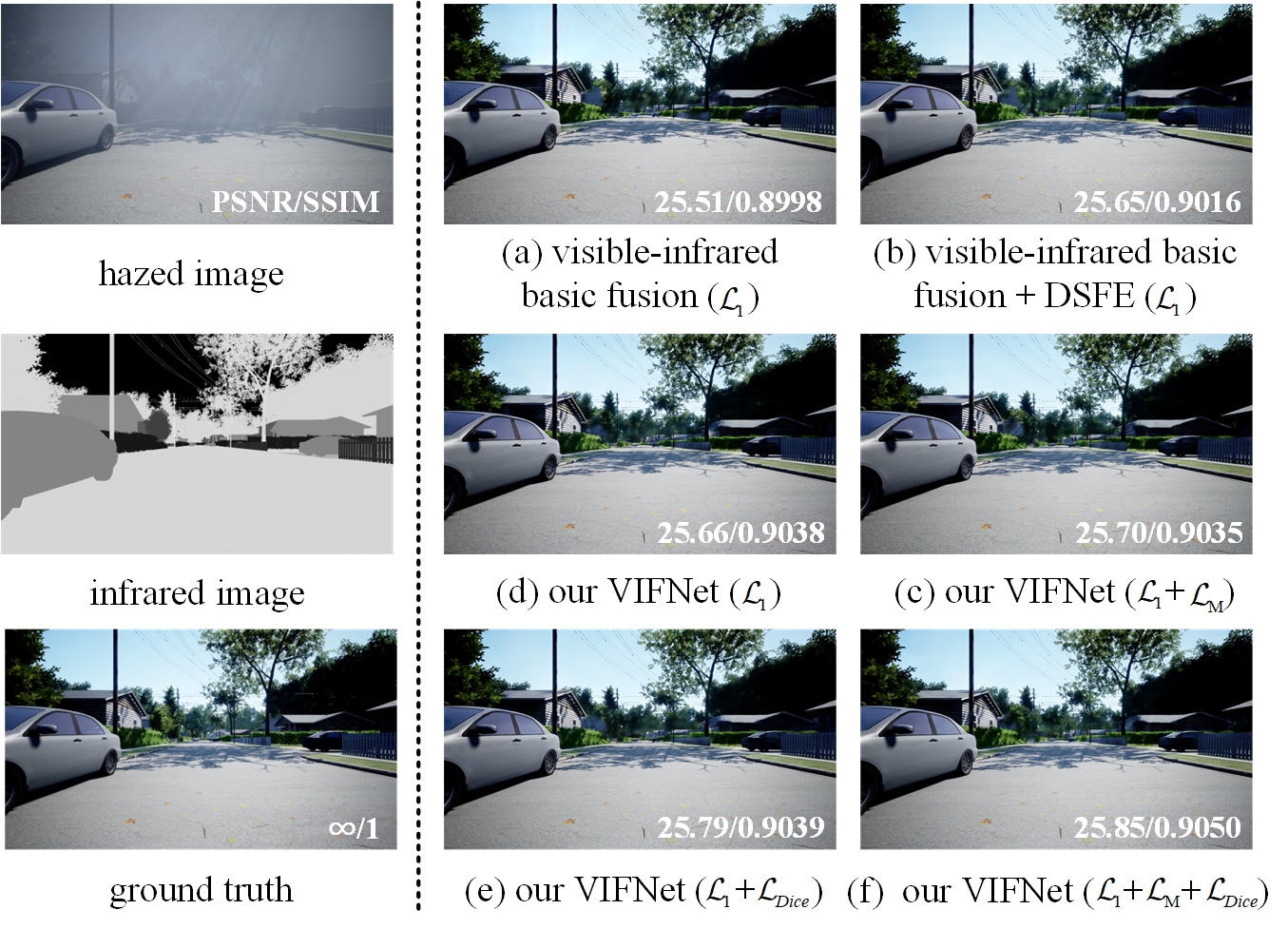}}
\caption{Quantitative and qualitative results of the ablation studies. The fog type of the input hazed image is medium haze.}
\label{fig10}
\end{figure}

In our experiment, we intentionally introduce misalignment between the two modalities in the AirSim-VID dataset to validate its impact on the model's performance. The misalignment was set at 30 pixels, which is relative to the size of 240$\times$240. Fig. 10 illustrates the dehazing performance of the model in the presence of misalignment. Although the haze can be removed to some extent, the non-aligned areas marked by the red boxes are not satisfactory and exhibit noticeable artifacts and inconsistencies. This issue could potentially be attributed to the fusion process of the two modalities.
}


\section{Conclusion and Perspectives}
\minew{This paper introduces the incorporation of infrared modality for image dehazing, and the proposed VIFNet achieves superior performance on various datasets. In summary, we investigate a deep structural feature fusion approach that combines visible and infrared modalities using an inconsistency fusion strategy. This approach effectively preserves crucial information and maximizes the benefits of each modality, which allows for the removal of dense haze areas when the visible image is blurred, in comparison to other methods.

However, the VIFNet still has certain limitations. Similar to most visible-infrared fusion methods, VIFNet is more suitable for scenes where there is strict alignment between the two modalities. In the future, we plan to address this issue by integrating alignment mechanisms into the network architecture.}


\begin{figure}[!ht]
\centerline{\includegraphics[width=\columnwidth]{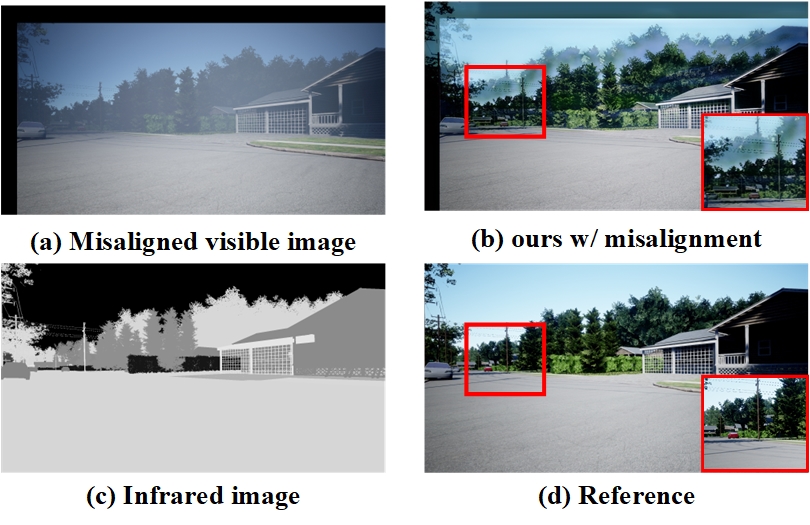}}
\caption{\minew{Dehazing results in the presence of misalignment between the two modalities.}}
\label{fig11}
\end{figure}



\bibliographystyle{cas-model2-names}

\bibliography{cas-refs}

\bio{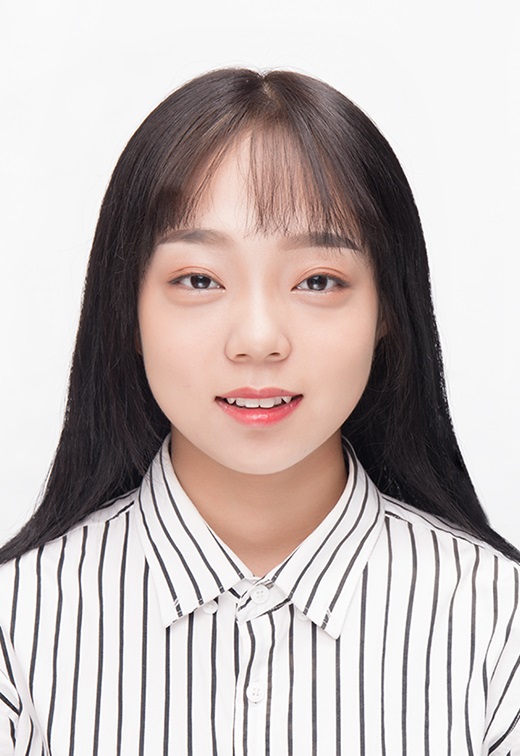}
\minew{\textbf{Meng Yu} received the B.Eng. degree in automation from the School of Instrument and Electronics, North University of China, in 2020. She is currently a Ph.D student in Control Science and Engineering with the School of Automation, Beijing Institute of Technology. Her research interests include multimodal sensor fusion, robotics perception, and computer vision.}
\endbio

\bio{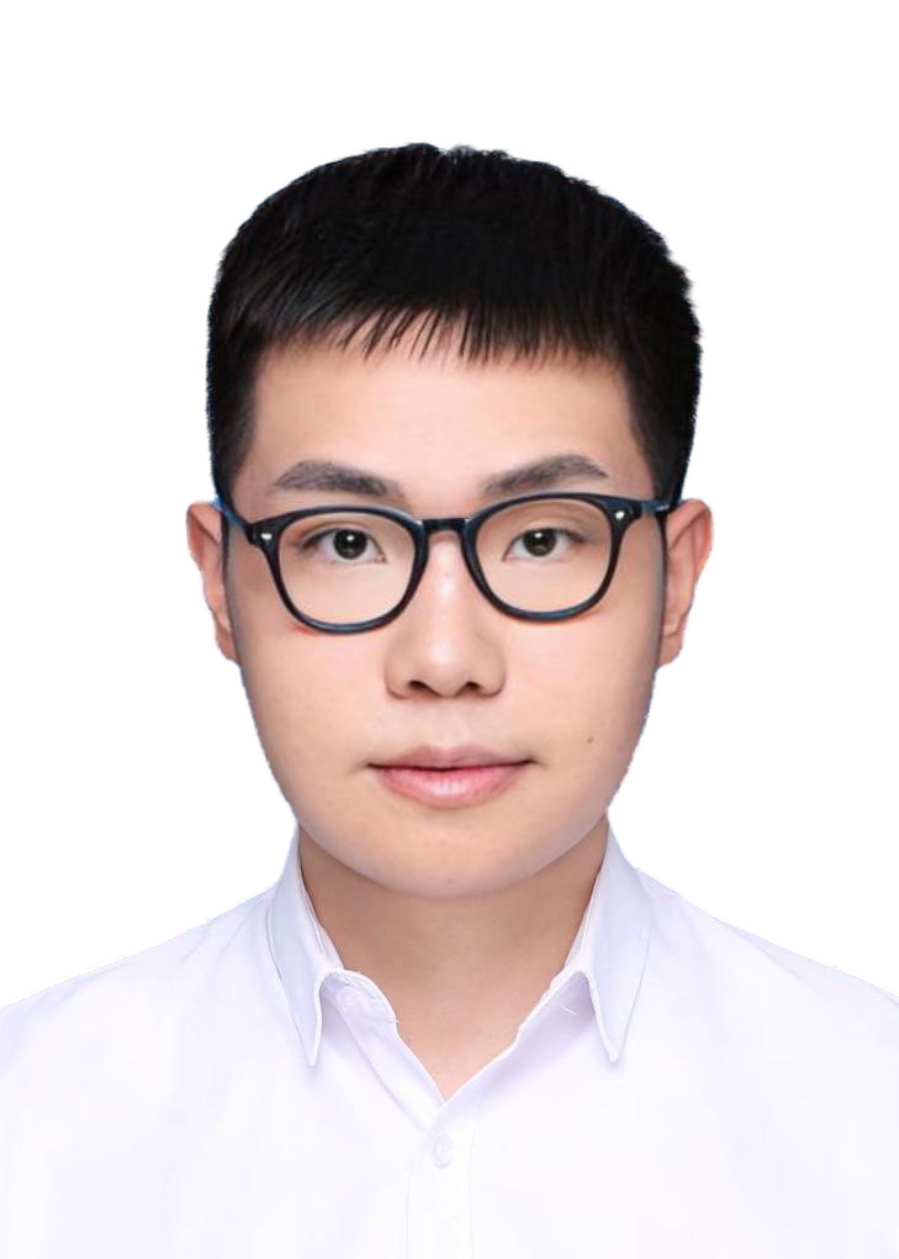}
\minew{\textbf{Te Cui} received the B.Eng. degree in automation from the School of Xuteli, Beijing Institute of Technology, in 2022. He is currently a Ph.D student in Control Science and Engineering with the School of Automation, Beijing Institute of Technology. His research interests include image feature matching and visual localization for robotics.}
\endbio

\bio{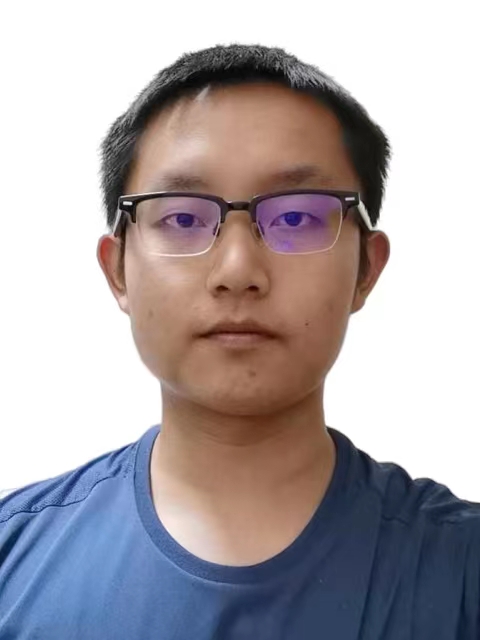}
\minew{\textbf{Haoyang Lu} received the B.Eng. degree in automation from the School of Automation, Beijing Institute of Technology, in 2023. He is currently a master student in Navigation, Guidance and Control with the School of Automation, Beijing Institute of Technology. His research interests include deep learning on point clouds and multi-sensor calibration.}
\endbio

\bio{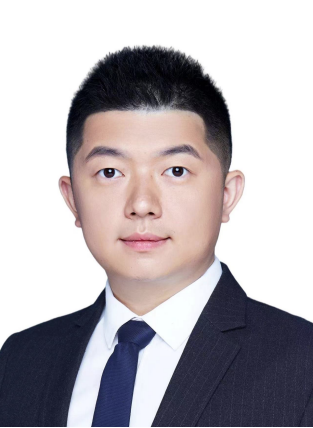}
\minew{\textbf{Yufeng Yue} (Member, IEEE) received the B.Eng. degree in automation from the Beijing Institute of Technology, Beijing, China, in 2014, and the Ph.D. degree in electrical and electronic engineering from Nanyang Technological University, Singapore, in 2019. He is currently a Professor with School of Automation, Beijing Institute of Technology. He has published a book in Springer, and more than 60 journal/conference papers, including IEEE TMM/TMech/TII/TITS, and conferences like NeurIPS/ICCV/ICRA/IROS. He is an Associate Editor for 2020–2023 IEEE IROS. His research interests include perception, mapping and navigation for autonomous robotics.}
\endbio
\end{document}